\def\year{2016}\relax
\documentclass[journal,transmag]{IEEEtran}

\usepackage{times}
\usepackage{helvet}
\usepackage{courier}

\usepackage{graphicx}
\usepackage{amsmath,amssymb} 
\usepackage{color}
\usepackage{times}
\usepackage{epsfig}
\usepackage{float}
\usepackage{url}
\usepackage{subfig}
\usepackage{multirow}
\usepackage{algorithm}
\usepackage{algpseudocode}
\usepackage{booktabs}

\usepackage{array}
\newcolumntype{x}[1]{>{\centering\arraybackslash\hspace{0pt}}m{#1}}
\usepackage{epstopdf}
\begin{document}

\title{Visual Question: Predicting If a Crowd Will Agree on the Answer} 

\author{
	\IEEEauthorblockN{Danna~Gurari and
		Kristen~Grauman}
	\IEEEauthorblockA{Department of Computer Science \\ The University of Texas at Austin}
}

\maketitle

\begin{abstract}
Visual question answering (VQA) systems are emerging from a desire to empower users to ask any natural language question about visual content and receive a valid answer in response.  However, close examination of the VQA problem reveals an unavoidable, entangled problem that multiple humans may or may not always agree on a single answer to a visual question.  We train a model to automatically predict from a visual question whether a crowd would agree on a single answer.  We then propose how to exploit this system in a novel application to efficiently allocate human effort to collect answers to visual questions.  Specifically, we propose a crowdsourcing system that automatically solicits fewer human responses when answer agreement is expected and more human responses when answer disagreement is expected.  Our system improves upon existing crowdsourcing systems, typically eliminating at least 20\% of human effort with no loss to the information collected from the crowd.  
\end{abstract}

\section{Introduction}
What would be possible if a person had an oracle that could immediately provide the answer to any question about the visual world?  Sight-impaired users could quickly and reliably figure out the denomination of their currency and so whether they spent the appropriate amount for a product~\cite{BighamJaJiLiMiMiMiTaWhWhYe10}.  Hikers could immediately learn about their bug bites and whether to seek out emergency medical care.  Pilots could learn how many birds are in their path to decide whether to change course and so avoid costly, life-threatening collisions.  These examples illustrate several of the interests from a \emph{visual question answering (VQA) system}, including tackling problems that involve classification, detection, and counting.  More generally, the goal for VQA is to have a \emph{single} system that can accurately answer any natural language question about an image or video~\cite{AndreasRoDaKl16,AntolAgLuMiBaZiPa15,MalinowskiFr14_2}.  

\begin{figure}[t!] 
\centering
\includegraphics[width=0.45\textwidth]{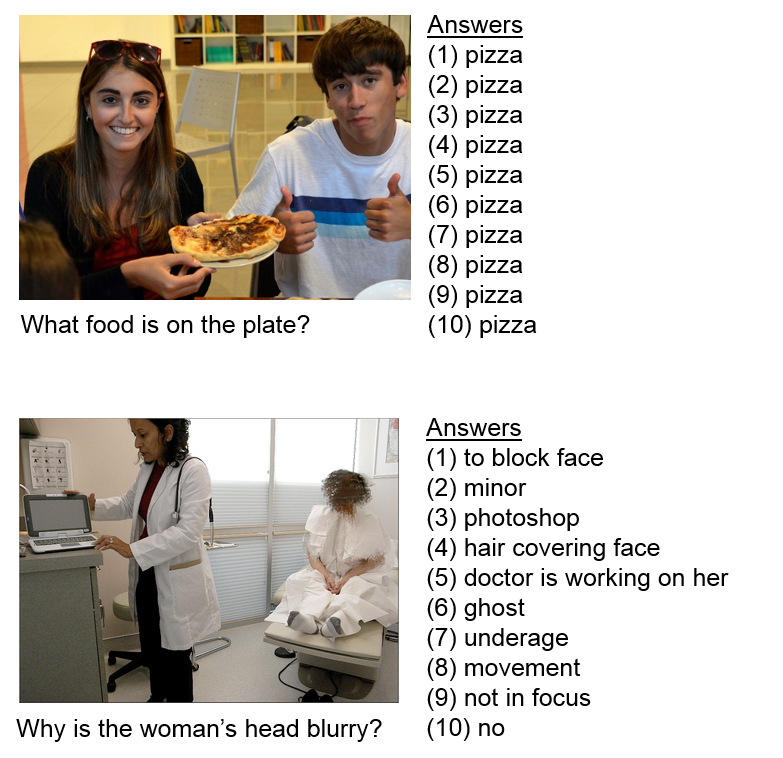}
\caption{Examples of visual questions and corresponding answers, when 10 different people are asked to answer the same question about an image (``visual question").  As observed, the crowd sometimes all agree on a single answer (top) and at other times offer different answers (bottom).  Given a visual question, we aim to build a prediction system that automatically decides whether multiple people would give the same answer.}
\label{fig_motivationPic}
\end{figure} 

Entangled in the dream of a VQA system is an unavoidable issue that, when asking multiple people a visual question, sometimes they all agree on a single answer while other times they offer different answers (\textbf{Figure~\ref{fig_motivationPic}}).  In fact, as we show in the paper, these two outcomes arise in approximately equal proportions in today's largest publicly-shared VQA benchmark that contains over 450,000 visual questions.  \textbf{Figure~\ref{fig_motivationPic}} illustrates that human disagreements arise for a variety of reasons including different descriptions of the same concept (e.g., ``minor" and ``underage"), different concepts (e.g., ``ghost" and ``photoshop"), and irrelevant responses (e.g., ``no").

Our goal is to \emph{account for whether different people would agree on a single answer to a visual question to improve upon today's VQA systems}.  We propose multiple prediction systems to automatically decide whether a visual question will lead to human agreement and demonstrate the value of these predictions for a new task of capturing the diversity of all plausible answers with less human effort. 

Our work is partially inspired by the goal to improve how to employ crowds as the computing power at run-time.  Towards satisfying existing users, gaining new users, and supporting a wide range of applications, a crowd-powered VQA system should be low cost, have fast response times, and yield high quality answers.  Today's status quo is to assume a fixed number of human responses per visual question and so a fixed cost, delay, and potential diversity of answers for every visual question~\cite{AntolAgLuMiBaZiPa15,BighamJaJiLiMiMiMiTaWhWhYe10,YuPaBeBe15}.  We instead propose to dynamically solicit the number of human responses based on each visual question.  In particular, we aim to accrue additional costs and delays from collecting extra answers only when extra responses are needed to discover all plausible answers.  We show in our experiments that our system saves 19 40-hour work weeks and \$1800 to answer 121,512 visual questions, compared to today's status quo approach~\cite{BighamJaJiLiMiMiMiTaWhWhYe10}.

Our work is also inspired by the goal to improve how to employ crowds to produce the information needed to train and evaluate automated methods.  Specifically, researchers in fields as diverse as computer vision~\cite{AntolAgLuMiBaZiPa15}, computational linguistics~\cite{AndreasRoDaKl16}, and machine learning~\cite{MalinowskiFr14_2} rely on large datasets to improve their VQA algorithms.  These datasets include visual questions and human-supplied answers.  Such data is critical for teaching machine learning algorithms how to answer questions by example.  Such data is also critical for evaluating how well VQA algorithms perform.  In general, ``bigger" data is better.  Current methods to create these datasets assume a fixed number of human answers per visual question~\cite{AntolAgLuMiBaZiPa15,YuPaBeBe15}, thereby either compromising on quality by not collecting all plausible answers or cost by collecting additional answers when they are redundant.  We offer an economical way to spend a human budget to collect answers from crowd workers.  In particular, we aim to actively allocate additional answers only to visual questions likely to have multiple answers.  

The key contributions of our work are as follows:

\begin{itemize}
\item Analysis demonstrating the prevalence and reasons for human answer disagreements in today's largest, freely-available VQA benchmark. 
\item A new problem and system for predicting whether a crowd will (dis)agree when answering a visual question.
\item A novel application for efficient answer collection which solicits additional answers from additional members of a crowd when disagreement is anticipated.
\end{itemize}

\section{Related Work}

\paragraph*{Visual Question Answering Services}
Researchers spanning communities as diverse as human computer interaction, machine learning, computational linguistics, and computer vision have proposed a variety of ways to answer questions about images~\cite{AndreasRoDaKl16,AntolAgLuMiBaZiPa15,BighamJaJiLiMiMiMiTaWhWhYe10,MalinowskiFr14_2}.  Yet a commonality across these communities is they adopt a one-size-fits-all approach when deciding the number of answers for any visual question.  For example, crowd-powered systems aim to supply a pre-specified, fixed number of answers per visual question~\cite{BighamJaJiLiMiMiMiTaWhWhYe10} and automated systems return a single answer for every visual question~\cite{AndreasRoDaKl16,AntolAgLuMiBaZiPa15,MalinowskiFr14_2}.  Inspired by the observation that there can be multiple plausible answers per visual question, we propose a richer representation of visual question answering that accounts for whether different people would agree on a single answer.  We propose a system that automatically predicts whether humans will disagree.  We demonstrate the predictive advantage of our system over relying on the uncertainty of a VQA algorithm in its predicted answer~\cite{AntolAgLuMiBaZiPa15}.

\paragraph*{Answer Collection from a Crowd} 
Our work relates to methods that propose how to employ crowd workers to answer questions about images.  Such approaches aim to collect a pre-specified, fixed number of answers per visual question.  For those systems that treat response time as a first priority, a variable number of answers may arise but this is due to varying crowdsourcing conditions such as the available supply of workers~\cite{BighamJaJiLiMiMiMiTaWhWhYe10,BurtonBrBrNeBiHu12}.  Other systems ensure a fixed number of answers are collected per visual question~\cite{AntolAgLuMiBaZiPa15,YuPaBeBe15}.  Unlike prior work, our goal is to collect answers in a way that is both economical \emph{and} complete in capturing the diversity of plausible answers for all visual questions.  To our knowledge, our work is the first to predict the number of answers to collect for a visual question.  Experiments demonstrate that our disagreement predictions are useful to significantly reduce human effort for capturing the diversity of valid answers for 121,512 visual questions.  

\paragraph*{Analyses of Crowd Disagreement}
More broadly, our work relates to efforts to account for crowd disagreement.  For example, researchers have suggested ways to resolve crowd disagreement due to task difficulty~\cite{WelinderBrBePe10} and ambiguity/specificity~\cite{JasPa15,AmidUk15}.  Some methods demonstrate which workers to trust most when aggregating multiple responses into a final, single response~\cite{SheshadriLease13,WelinderBrBePe10}.  Other methods leverage context to automatically disambiguate which of multiple outcomes is the desired outcome~\cite{AmidUk15}.  Unlike prior work, we focus on the task of visual question answering.  Moreover, while prior work focuses on resolving specific sources of crowd disagreement (e.g., task difficulty or ambiguity), we instead propose a single, integrated system that jointly detects various sources of crowd disagreement that arise for visual question answering.  The advantage of this approach is to separate ``easy to answer" instances from all instances that would require additional effort to resolve the disagreement; e.g., collect multiple answers for ambiguous and subjective tasks or apply an aggregation scheme to produce a single answer from multiple answers when crowd workers are unreliable.

\paragraph*{High Quality Work with Fixed Human Budget}  
Our work aligns with methods that actively allocate a limited human budget to where it will best contribute to improving the quality of results.  For example, one method distributes a budget between three different levels of human effort when deciding how to segment images~\cite{JainGr13}.  Another method spends a budget between less costly crowd workers and more costly expert efforts to improve outcomes for biomedical citation screening~\cite{NguyenWaLe15}.  Another method predicts when to employ algorithms versus crowd workers to segment images~\cite{GurariJaBeGr16}.  To our knowledge, our work is the first towards deciding how to spend a budget for the task of visual question answering, which is distinct from prior work which focused on spending a budget for image analysis or language analysis alone.  Furthermore, our aim is to spend a budget to capture the \emph{diversity} of all valid results for every task rather than to collect a \emph{single} result for every task.

\paragraph*{Minimizing Human Labeling}  
Our aim to actively decide how to allocate human effort to improve results is also somewhat related to active learning~\cite{Settles10}.  Specifically, active learners try to use as little human effort as possible to train accurate prediction models.  Some methods iteratively supplement a training dataset with the most informative images for training a classifier~\cite{PattersonHoBePeHa15,VijayanarasimhanGr11}.  Other methods solicit redundant labels to prevent incorrect/noisy labels from teaching prediction models to make mistakes~\cite{LinMaWe14,ShengPrIp08}.  While active learners aim to minimize human input to improve the accuracy of a prediction model, our method aims to minimize human input while still exhaustively capturing all plausible answers to all visual questions.

\paragraph*{Continuous Dialogue with the Crowd} 
Two services - Be My Eyes~\cite{BeMyEyes16} and Chorus:View~\cite{LaseckiThZhBrBi13} - offer users a continuous communication channel with members of the crowd to answer visual questions.  The aim is to expedite arriving at desired answers to, for example, clarify ambiguous questions.  Our work offers an alternative by demonstrating how a crowdsourcing service might instead solicit multiple answers for a one time back-and-forth rather than enacting a more costly, continuous communication channel with a single voice, whether from a single person~\cite{BeMyEyes16} or the consensus of a crowd~\cite{LaseckiThZhBrBi13}.  

\section{Paper Overview}
The remainder of the paper is organized into four sections.  We first describe a study where we investigate: 1) How much answer diversity arises for visual questions? and 2) Why do people disagree (\textbf{Section~\ref{sec_disagreementAnalysis}})?  Next, we explore the following two questions: 1) Given a novel visual question, can a machine correctly predict whether multiple independent members of a crowd would supply the same answer? and 2) If so, what insights does our machine-learned system reveal regarding what humans are most likely to agree about (\textbf{Section~\ref{sec_predictingDisagreement}})?  In the following section, we propose a novel resource allocation system for efficiently capturing the diversity of all answers for a set of visual questions (\textbf{Section~\ref{sec_resourceAllocation}}).  Finally, we end with concluding remarks (\textbf{Section~\ref{sec_conclusions}}).    

\section{VQA - Analysis of Answer (Dis)Agreements}
\label{sec_disagreementAnalysis}
Our first aim is to answer the following questions: 1) How much answer diversity arises for visual questions? and 2) Why do people disagree?   

\paragraph*{VQA Datasets}
We conduct our analysis on a total of 459,861 visual questions and 4,598,610 answers coming from today's largest freely-available VQA benchmark~\cite{AntolAgLuMiBaZiPa15}.  We chose this benchmark because it both represents a diversity of visual questions and includes many crowdsourced answers for every visual question.  

The benchmark consists of two datasets that reflect VQAs for \emph{real images} and \emph{abstract scenes}.  Specifically,
80\% (i.e., 369,861) of VQAs are about real images that show 91 types of objects that would be ``easily recognizable by a 4 year old" in their natural context~\cite{LinMaBeHaPeRaDoZi14}.  The remaining 90,000 VQAs are about abstract scenes that were created with clipart and show 100 types of everyday objects often observed in real images~\cite{AntolAgLuMiBaZiPa15}.    

The benchmark includes a diversity of \emph{visual questions} intentionally collected to be both grounded in images and task-independent.  Towards this aim, visual questions were collected by asking three Amazon Mechanical Turk (AMT) crowd workers to look at a given image and generate a text-based question about it that would  ``stump a smart robot"~\cite{AntolAgLuMiBaZiPa15}.  Three open-ended questions were collected about each of 153,287 images, resulting in a total of 459,861 visual questions.    
  
The benchmark also includes 10 open-ended \emph{natural language answers} from 10 AMT crowd workers per visual question.  Each answer was collected by showing a worker an image with associated question and asking him/her to respond with ``a brief phrase and not a complete sentence"~\cite{AntolAgLuMiBaZiPa15}.  

Finally, to enrich our analysis, we leverage the included labels which indicate the type of answer elicited for each visual question.  Specifically, each visual question is labeled as eliciting one of the following types of answers: ``yes/no", ``number", or ``other".  This label was determined as the most popular option from 10 labels assigned to the associated 10 answers for each visual question.    

\paragraph*{Defining Answer Diversity}
We compute answer diversity for a visual question by counting how many \emph{unique} and \emph{valid} answers are observed in the set of answers.  We derive results using the 10 crowdsourced answers per visual question.  

We establish \emph{unique answers} by pre-processing each answer to eliminate cosmetic differences and then applying exact string matching to identify the number of different answers.  We pre-process each answer by converting all letters to lower case, converting numbers to digits, and removing punctuation and articles (i.e., ``a", ``an", ``the"), as was done in prior work~\cite{AntolAgLuMiBaZiPa15}.  While this approach does not fully resolve all conceptually equivalent responses, it does reveal an upper bound of expected answer diversity.  In other words, more lenient agreement schemes (e.g., employing sophisticated natural language processing methods) would lead to either the same or less answer diversity.

We establish \emph{valid answers} by tallying the number of times each unique answer is given and then only accepting answers observed from at least $m$ people, where $m$ is an application-specific parameter to set.  In practice, prior work deems answers as 100\% valid using blind trust (i.e., $m=1$ person)~\cite{MalinowskiRoFr15} as well as more conservative answer validation schemes (i.e., $m=3$ people)~\cite{AntolAgLuMiBaZiPa15}.  

\begin{figure}[!t]
\centering
\includegraphics[width=0.41\textwidth]{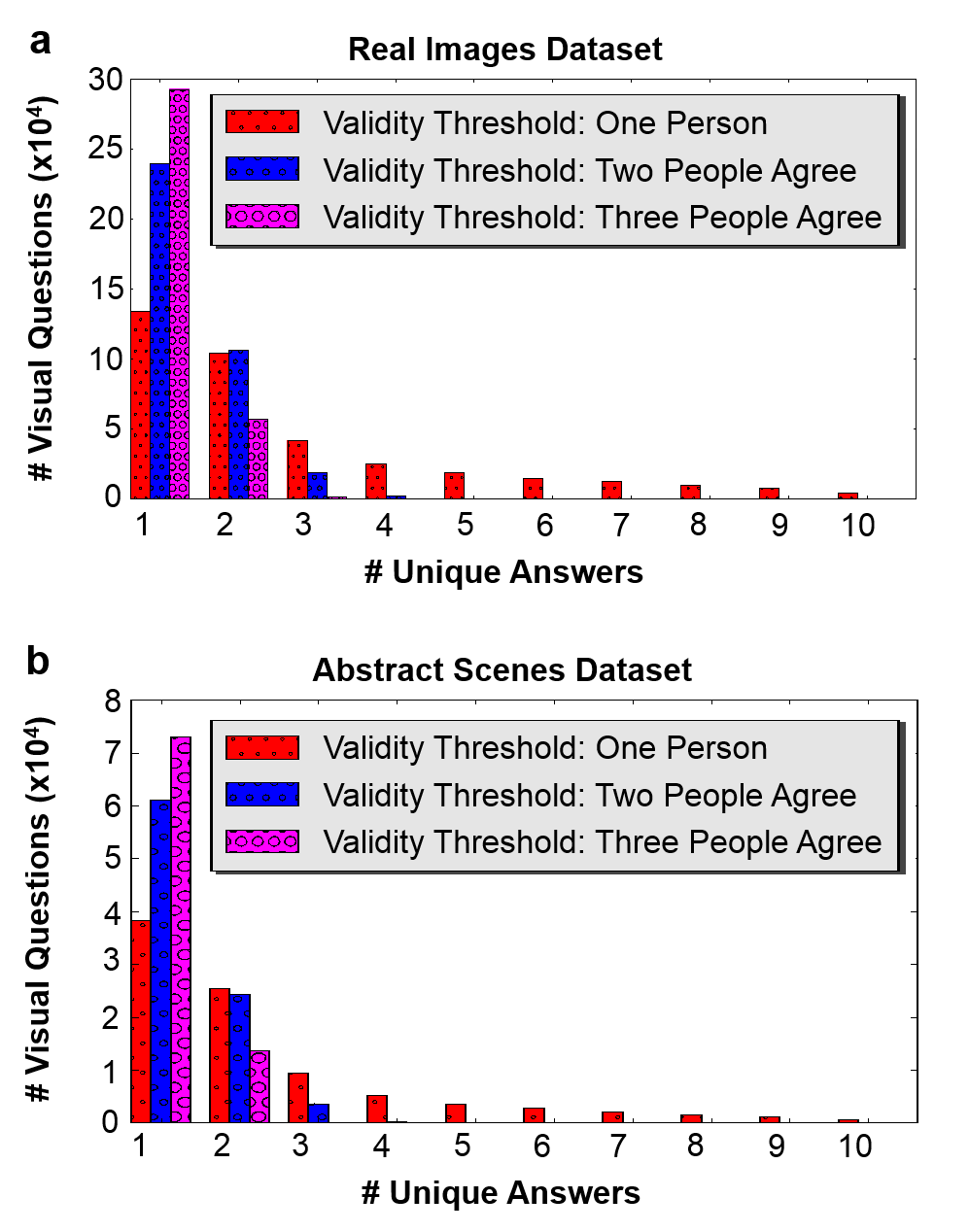}
\vspace{-0.5em}
  \caption{Summary of answer diversity outcomes showing how frequently different numbers of unique answers arise when asking ten crowd workers to answer a visual question for (\textbf{a}) 369,861 visual questions about real images and (\textbf{a}) 90,000 visual questions about abstract scenes.  Results are shown based on different degrees of answer agreement required to make an answer valid: only one person has to offer the answer, at least two people must agree on the answer, and at least three people must agree on the answer.  The visual questions most often elicit exactly one answer per visual question but also regularly elicit up to three different answers per visual question.}
\label{fig_answerDiversityQuantified}
\end{figure}

\paragraph*{Measuring Answer Diversity}
We now turn to the question of how much answer \emph{diversity} is observed in practice for visual questions.  Across all 459,861 visual questions, we tally how many visual questions yield $k$ unique, valid answers where $k = \{1, 2, ..., 10\}$.   

We enrich our analysis on measured answer diversity by examining the influence of different levels of trust in the crowd as well as the influence of different datasets.  Specifically, we tally the number of unique, valid answers observed when requiring a minimum of $m = 1$, $m = 2$, or $m = 3$ members of the crowd to offer the same answer for an answer to be valid.  We conduct our analysis on the two datasets of visual questions about real images and abstract scenes independently.    

The majority of visual questions lead to at most three unique answers, across all crowd trust levels and both datasets (\textbf{Figure~\ref{fig_answerDiversityQuantified}}).  This gives an upper bound of expected answer diversity.  We anticipate measured answer diversity will drop with less stringent answer agreement schemes than exact string matching, such as inferring agreement when an answer is a synonym or plurality of another answer.  

Our results show the same trend for the amount of answer diversity with all three agreement thresholds, for both datasets (\textbf{Figure~\ref{fig_answerDiversityQuantified}}).  Most commonly there is one unique answer, followed by two and three answers respectively.  In addition, as expected, moving from requiring no answer agreement to a more conservative agreement between three people shifts the overall distribution to more sharply peak at less overall diversity (i.e., 1 unique answer).  Our results also show that VQA statistics can transfer from one source of images to another, revealing a possible benefit of using artificially-generated images to learn trends when more costly real world images are not readily-available (\textbf{Figure~\ref{fig_answerDiversityQuantified}a} vs \textbf{\ref{fig_answerDiversityQuantified}b}).    

\begin{table*}[t!]
  \centering
        \begin{tabular}{ l  c  c  c | c  c  c }
    \toprule
      & \multicolumn{3}{c}{{\bf Real Images - 369,861 VQAs }} & \multicolumn{3}{c}{{\bf Abstract Scenes - 90,000 VQAs}}  \\ 
    \cmidrule(r){2-4} \cmidrule(r){5-7} 
     \textbf{Answer Type:}
    & {\small \textit{Yes/No}}
    & {\small \textit{Number}}
    & {\small \textit{Other}}
    & {\small \textit{Yes/No}}
    & {\small \textit{Number}}
    & {\small \textit{Other}} \\
     \textbf{\# VQAs (\%):}
    & {\small \textit{140,777 (38\%)}}
    & {\small \textit{45,822 (12\%)}}
    & {\small \textit{183,262 (50\%)}}
    & {\small \textit{36,717 (41\%)}}
    & {\small \textit{12,956 (14\%)}}
    & {\small \textit{40,327 (45\%)}} \\
    \midrule
       \textbf{At Most One Disagreement} & \textbf{74\%} & \textbf{49\%} & \textbf{35\%} & \textbf{74\%} & \textbf{79\%} & \textbf{36\%} \\ 
       \textbf{- Unanimous Agreement} & 54\% & 35\% & 22\% & 57\% & 65\% & 22\% \\ 
       \textbf{- Exactly One Disagreement} & 20\% & 14\% & 13\% & 17\% & 14\% & 14\% \\ 
       \bottomrule	 
  \end{tabular}
        \caption{Correlation between answer agreement and visual questions that elicit three different types of answers for VQAs on real images and abstract scenes.  Shown for each answer type is the percentage of visual questions that lead to at most one disagreement (row 1), unanimous agreement (row 2), and exactly one disagreement (row 3) from 10 crowdsourced answers.  On average, across all answer types for both datasets, the crowd agrees on the answer for nearly half (i.e., 53\%) of all VQAs.  Moreover, we observe crowd disagreement arises often for all three answer types, highlighting that the significance of crowd disagreement is applicable across various types of visual questions.}
        ~\label{table_answerDistributionAnalysis}
\end{table*} 

From the 369,861 visual questions about real images, we found that only 1\% (i.e., 3,992) and 5\% (i.e., 19,682) of visual questions have no valid answer when limiting valid answers to those which have at least two or three people agreeing on them respectively.  This suggests that a crowd is able to reach some level of consensus on what are acceptable answers for the vast majority of visual questions.  

\begin{figure*}[t!] 
\centering
\includegraphics[width=1\textwidth]{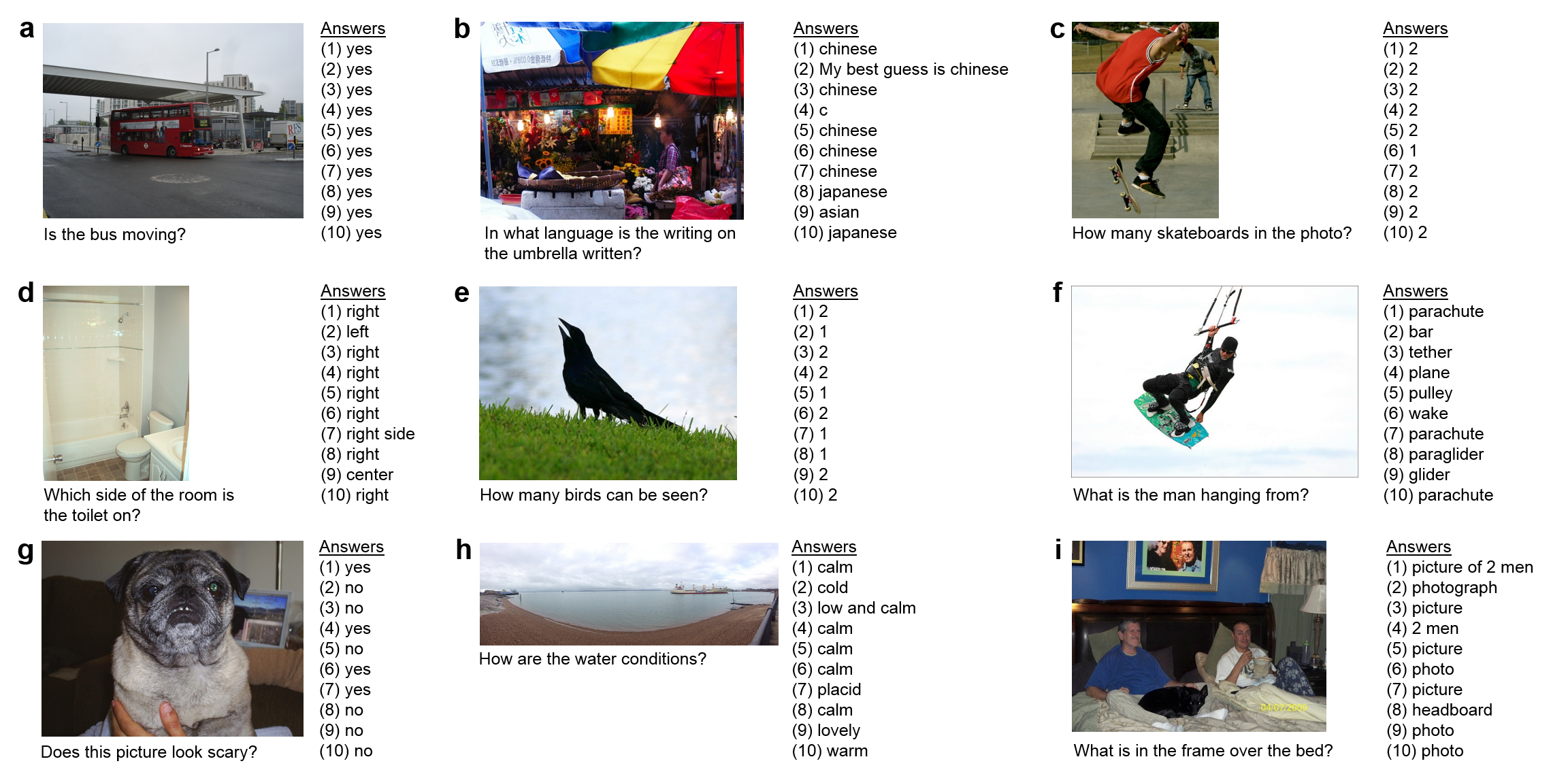}
  \vspace{-1.5em}
\caption{Illustration of visual questions that lead humans to (dis)agree on a single answer.  As observed, (\textbf{a}) unanimous answer agreement arises when images are simple, questions are precise, and questions are visually grounded.  (\textbf{b-i}) Answer disagreement arises for a variety of reasons: (\textbf{b}) expert skill needed, (\textbf{c}) human mistakes, (\textbf{d}) ambiguous question, (\textbf{e}) ambiguous visual content, (\textbf{f}) insufficient visual evidence, (\textbf{g}) subjective question, (\textbf{h}) answer synonyms, and (\textbf{i}) varying answer granularity.}
\label{fig_convAndDivExamples}
\end{figure*}  

\paragraph*{Reasons for Answer (Dis)Agreements}
Our second aim in analyzing the VQA benchmark is to better understand why people disagree when answering visual questions.  

\textbf{Figure~\ref{fig_convAndDivExamples}} highlights various reasons for why crowd workers disagree on an answer.  Disagreements can arise due to crowd worker skill, both because a difficult task necessitates domain expertise and because a crowd worker may inadequately answer a seemingly simple question (\textbf{Figure~\ref{fig_convAndDivExamples}b,c}).  Crowds disagree also because of ambiguity in the question and visual content (\textbf{Figure~\ref{fig_convAndDivExamples}d,e}).  Further reasons for disagreement include insufficient visual evidence to answer the question, subjective questions, synonymous answers, and varying levels of answer granularity (\textbf{Figure~\ref{fig_convAndDivExamples}f--i}).  We capitalize on these observations in the next section to design prediction systems that automatically separate visual questions that lead to agreement.  

We enrich our understanding of why crowds disagree by examining how frequently crowds (dis)agree with respect to visual questions that elicit different types of answers.  We tally the number of visual questions that lead to ``yes/no", ``number", and ``other" answers.  We report results for both when crowds unanimously agree as well as when nine of the ten people agree for both datasets (\textbf{Table~\ref{table_answerDistributionAnalysis}}).  These results capture when at most one untrusted result is permitted from the crowd when inferring whether a crowd agrees.  Overall, we observe at most one disagreement for 51.6\% of real images and 57.6\% for abstract scenes.  We find that disagreement arises often for all types of answers, highlighting that the interest in crowd disagreement is of widespread interest for many types of visual questions and different datasets.   

We observe similar crowd agreement trends across the two datasets for two of the three answer types (\textbf{Table~\ref{table_answerDistributionAnalysis}}).  We find high agreement for ``yes/no" images for both datasets.  We hypothesize that the remaining quarter of asked ``yes/no" questions that lead to greater amounts of disagreement are subjective questions and so lead to split opinions among a crowd (e.g., ``Does this picture look scary?", \textbf{Figure~\ref{fig_convAndDivExamples}g}).  We observe moderate agreement levels for ``other" visual questions, possibly due to a greater diversity of opinions regarding the true answer as well as ways to express the same concept.  We find the greatest difference between results for the two datasets on ``number" visual questions.  We hypothesize counting problems are easier for less complex images that show few objects, as is consistently the case for the abstract scenes but not the real images.  

\section{Visual Question - Predicting If a Crowd Will (Dis)Agree on an Answer}
\label{sec_predictingDisagreement}
We now explore the following two questions: 1) Given a novel visual question, can a machine correctly predict whether multiple independent members of a crowd would supply the same answer? and 2) If so, what insights does our machine-learned system reveal regarding what humans are most likely to agree about?  

\subsection*{Prediction Systems}
\label{sec_predictionSystem}
We pose the prediction task as a binary classification problem.  Specifically, given an image and associated question, a system outputs a binary label indicating whether a crowd will agree on the same answer.  Our goal is to design a system that can detect which visual questions to assign a disagreement label, regardless of the disagreement cause (e.g., subjectivity, ambiguity, difficulty).  We implement both random forest and deep learning classifiers.

\paragraph*{Answer (Dis)Agreement Labels} A visual question is assigned either an answer \emph{agreement} or \emph{disagreement} label.  To assign labels, we employ 10 crowdsourced answers for each visual question.  A visual question is assigned an answer \emph{agreement} label when there is an exact string match for 9 of the 10 crowdsourced answers (after answer pre-preprocessing, as discussed in the previous section) and an answer \emph{disagreement} label otherwise.  Our rationale is to permit the possibility of up to one ``careless/spam" answer per visual question.  The outcome of our labeling scheme is that a disagreement label is agnostic to the specific cause of disagreement and rather represents the many causes (described above).      

\paragraph*{Random Forest System} 
For our first system, we use domain knowledge to guide the learning process.  We compile a set of features that we hypothesize inform whether a crowd will arrive at an undisputed, single answer.  Then we apply a machine learning tool to reveal the significance of each feature.  We propose features based on the observation that answer agreement often arises when 1) a lay person's attention can be easily concentrated to a single, undisputed region in an image and 2) a lay person would find the requested task easy to address.  

\begin{figure*}[t!] 
\centering
\includegraphics[width=1\textwidth]{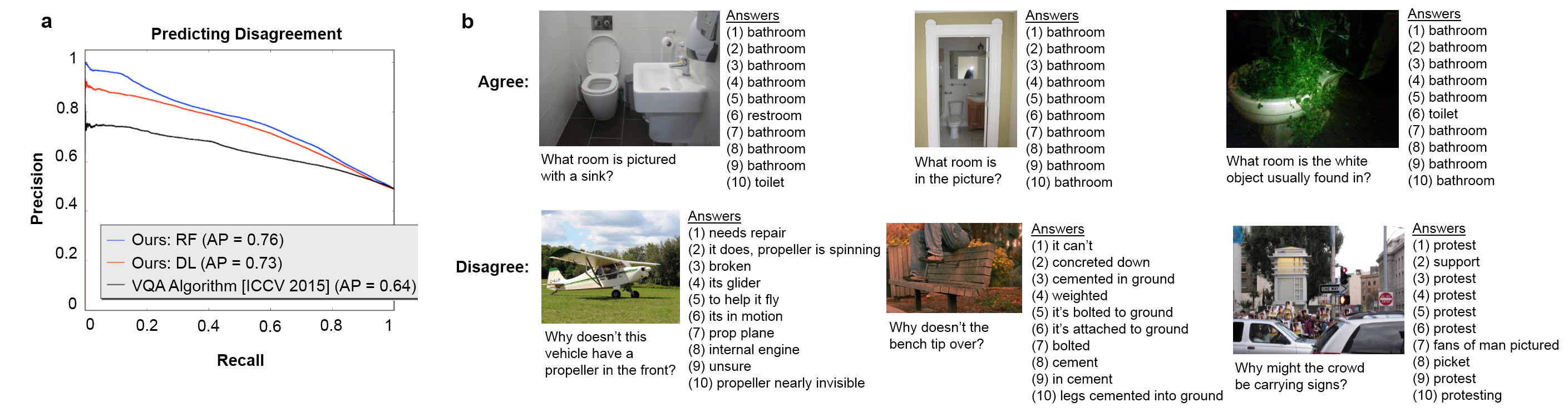}
\caption{(\textbf{a}) Precision-recall curves and average precision (AP) scores for all benchmarked systems.  Our random forest (RF) and deep learning (DL) classifiers outperform a related automated VQA baseline, showing the importance in modeling human disagreement as opposed to system uncertainty.  (\textbf{b}) Examples of prediction results from our top-performing RF classifier.  Shown are the top three visual questions for the most confidently predicted instances that lead to answer disagreement and agreement along with the observed crowd answers.  These examples illustrate a strong language prior for making predictions.  (Best viewed on pdf.)}
\label{fig_predictorAnalysis}
\end{figure*} 

We employ five \emph{image-based features} coming from the salient object subitizing~\cite{ZhangMaSaScBeLiShPrMe15} (SOS) method, which produces five probabilities that indicate whether an image contains 0, 1, 2, 3, or 4+ salient objects.  Intuitively, the number of salient objects shows how many regions in an image are competing for an observer's attention, and so may correlate with the ease in identifying a region of interest.  Moreover, we hypothesize this feature will capture our observation from the previous study that counting problems typically leads to disagreement for images showing many objects, and agreement otherwise.

We employ a 2,492-dimensional feature vector to represent the \emph{question-based features}.  One feature is the number of words in the question.  Intuitively, a longer question offers more information and we hypothesize additional information makes a question more precise.  The remaining features come from two one-hot vectors describing each of the first two words in the question.  Each one-hot vector is created using the learned vocabularies that define all possible words at the first and second word location of a question respectively (using training data, as described in the next section).  Intuitively, early words in a question inform the type of answers that might be possible and, in turn, possible reasons/frequency for answer disagreement.  For example, we expect ``why is" to regularly elicit many opinions and so disagreement.  This intuition about the beginning words of a question is also supported by our analysis in the previous section which shows that different answer types yield different biases of eliciting answer agreement versus disagreement.

We leverage a random forest \emph{classification model}~\cite{Breiman01} to predict an answer (dis)agreement label for a given visual question.  This model consists of an ensemble of decision tree classifiers.  We train the system to learn the unique weighted combinations of the aforementioned 2,497 features that each decision tree applies to make a prediction.  At test time, given a novel visual question, the trained system converts a 2,497 feature descriptor of the visual question into a final prediction that reflects the majority vote prediction from the ensemble of decision trees.  The system returns the final prediction along with a probability indicating the system's confidence in that prediction.  We employ the Matlab implementation of random forests, using 25 trees and the default parameters.     

\paragraph*{Deep Learning System} 
We next adapt a VQA deep learning architecture~\cite{LuLiBaPa15} to learn the predictive combination of visual and textual features.  The question is encoded with a 1024-dimensional LSTM model that takes in a one-hot descriptor of each word in the question.  The image is described with the 4096-dimensional output from the last fully connected layer of the Convolutional Neural Network (CNN), VGG16~\cite{SimonyanZi14}.  The system performs an element-wise multiplication of the image and question features, after linearly transforming the image descriptor to 1024 dimensions.  The final layer of the architecture is a softmax layer.  

We train the system to predict (dis)agreement labels with training examples, where each example includes an image and question.  At test time, given a novel visual question, the system outputs an unnormalized log probability indicating its belief in both the agreement and disagreement label.  For our system's prediction, we convert the belief in the disagreement label into a normalized probability.  Consequently, predicted values range from 0 to 1 with lower values reflecting greater likelihood for crowd agreement.    

\subsection*{Analysis of Prediction System}
\label{sec_predictionSystemEvaluation}
We now describe our studies to assess the predictive power of our classification systems to decide whether visual questions will lead to answer (dis)agreement.  

We capitalize on today's largest visual question answering dataset~\cite{AntolAgLuMiBaZiPa15} to evaluate our prediction system, which includes 369,861 visual questions about real images.  Of these, 248,349 visual questions (i.e., Training questions 2015 v1.0) are kept for for training  and the remaining 121,512 visual questions (i.e., Validation questions 2015 v1.0) are employed for testing our classification system.  This separation of training and testing samples enables us to estimate how well a classifier will generalize when applied to an unseen, independent set of visual questions.

To our knowledge, no prior work has directly addressed predicting answer (dis)agreement for visual questions.  Therefore, we employ as a baseline a related VQA algorithm~\cite{LuLiBaPa15,AntolAgLuMiBaZiPa15} which produces for a given visual question an answer with a confidence score.  This system parallels the deep learning architecture we adapt.  However, it predicts the system's uncertainty in its own answer, whereas we are interested in the humans' collective disagreement on the answer.  Still, it is a useful baseline to see if an existing algorithm could serve our purpose.  

\begin{figure*}[t!] 
\centering
\includegraphics[width=1\textwidth]{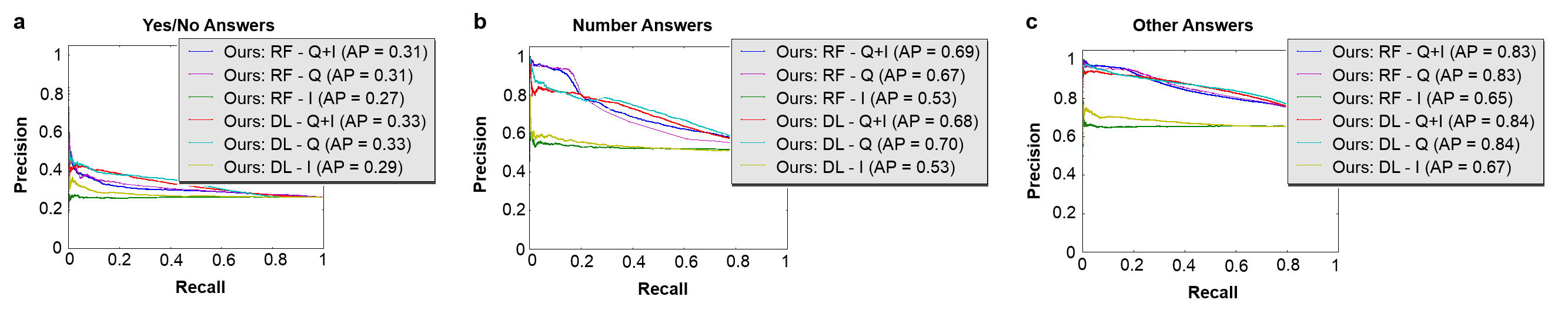}
\caption{Precision-recall curves and average precision (AP) scores for our random forest (RF) and deep learning (DL) classifiers with different features (Question Only - Q; Image Only - I; Q +I) for visual questions that lead to (\textbf{a-c}) three answer types.}
\label{fig_predictiveFeatureAnalysis}
\end{figure*} 

\paragraph*{Classification Performance} 
We evaluate the predictive power of the classification systems based on each classifier's predictions for the 121,512 visual questions in the test dataset.  We first show performance of the baseline and our two prediction systems using \emph{precision-recall curves}.  The goals are to achieve a high precision, to minimize wasting crowd effort when their efforts will be redundant, and a high recall, to avoid missing out on collecting the diversity of accepted answers from a crowd.  We also report the \emph{average precision} (AP), which indicates the area under a precision-recall curve.  AP values range from 0 to 1 with better-performing prediction systems having larger values.  

\textbf{Figure~\ref{fig_predictorAnalysis}a} shows precision-recall curves for all prediction systems.  Both our proposed classification systems outperform the \texttt{VQA Algorithm~\cite{AntolAgLuMiBaZiPa15}} baseline; e.g., \texttt{Ours - RF} yields a 12 percentage point improvement with respect to AP.  This is interesting because it shows there is value in learning the disagreement task specifically, rather than employing an algorithm's confidence in its answers.  More generally, our results demonstrate it is possible to predict whether a crowd will agree on a single answer from a given image and associated question.  Despite the significant variety of questions and image content, and despite the variety of reasons for which the crowd can disagree, our learned model is able to produce quite accurate results\footnote{We observe no change to predictive performance of the random forest classifier when instead training the model such that an agreement label is assigned to a visual question only when all 10 answers match.  In other words, we see no difference in predictive power when we flip labels for all examples where nine people agree and one person disagrees.}.

We observe our Random Forest classifier outperforms our deep learning classifier; e.g., \texttt{Ours: RF} yields a three percentage point improvement with respect to AP while consistently yielding improved precision-recall values over \texttt{Ours: LSTM-CNN} (\textbf{Figure~\ref{fig_predictorAnalysis}a}).  In general, deep learning systems hold promise to replace handcrafted features to pick out the discriminative features.  Our baselines highlight a possible value in developing a different deep learning architecture for the problem of learning answer disagreement than applied for predicting answers to visual questions.

We show examples of prediction results where our top-performing \texttt{RF} classifier makes its most confident predictions (\textbf{Figure~\ref{fig_predictorAnalysis}b}).  In these examples, the predictor expects human agreement for ``what room... ?" visual questions and disagreement for ``why... ?" visual questions.  These examples highlight that the classifier may have a strong language prior towards making predictions, as we will discuss in the next section.

\paragraph*{Predictive Cues}
We now explore what makes a visual question lead to crowd answer agreement versus disagreement.  We examine the influence of whether visual questions lead to the three types of answers (``yes/no", ``number", ``other") for both our random forest (\texttt{RF}) and deep learning (\texttt{DL}) classification systems.  We enrich our analysis by examining the predictive performance of both classifiers when they are trained and tested exclusively with image and question features respectively.  \textbf{Figure~\ref{fig_predictiveFeatureAnalysis}} shows precision-recall curves for both classification systems with question features alone (\texttt{Q}), image features alone (\texttt{I}), and both question and image features together (\texttt{Q+I}).  

When comparing AP scores (\textbf{Figure~\ref{fig_predictiveFeatureAnalysis}}), we observe our \texttt{Q+I} predictors yield the greatest predictive performance for visual questions that lead to ``other" answers, followed by ``number" answers, and finally ``yes/no" answers.  One possible reason for this finding is that the question wording strongly drives whether a crowd will disagree for ``other" visual questions, whereas some notion of common sense may be required to learn whether a crowd will agree for ``yes/no" visual questions (e.g., \textbf{Figure~\ref{fig_convAndDivExamples}a} vs \textbf{Figure~\ref{fig_convAndDivExamples}g}).  

We observe that question-based features yield greater predictive performance than image-based features for all visual questions, when comparing AP scores for \texttt{Q} and \texttt{I} classification results (\textbf{Figure~\ref{fig_predictiveFeatureAnalysis}}).  In fact, image features contribute to performance improvements only for our random forest classifier for visual questions that lead to ``number" answers, as illustrated by comparing AP scores for \texttt{Our RF: Q+I} and \texttt{Our RF: Q} (\textbf{Figure~\ref{fig_predictiveFeatureAnalysis}b}).  Our overall finding that most of the predictive power stems from language-based features parallels feature analysis findings in the automated VQA literature~\cite{AntolAgLuMiBaZiPa15,MalinowskiRoFr15}.  This does not mean, however, that the image content is not predictive.  Further work improving visual content cues for VQA agreement is warranted.  

\begin{figure*}[t!] 
\centering
\includegraphics[width=1\textwidth]{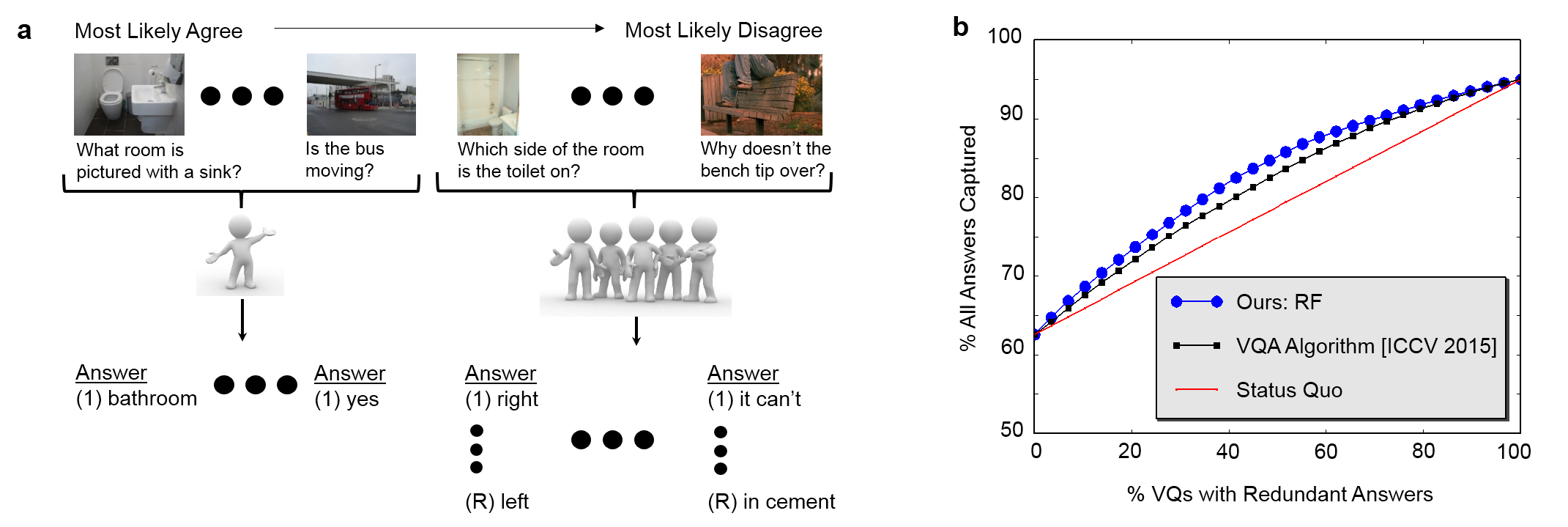}
\caption{We propose a novel application of predicting the number of redundant answers to collect from the crowd per visual question to efficiently capture the diversity of all answers for all visual questions.  (\textbf{a}) For a batch of visual questions, our system first produces a relative ordering using the predicted confidence in whether a crowd would agree on an answer (upper half).  Then, the system allocates a minimum number of annotations to all visual questions (bottom, left half) and then the extra available human budget to visual questions most confidently predicted to lead to crowd disagreement (bottom, right half). (\textbf{b}) For 121,512 visual questions, we show results for our system, a related VQA algorithm, and today's status quo of random predictions.  Boundary conditions are one answer (leftmost) and five answers (rightmost) for all visual questions.  Our approach typically accelerates the capture of answer diversity by over 20\% from today's \texttt{Status Quo} selection; e.g., 21\% for 70\% of the answer diversity and 23\% for 86\% of the answer diversity.  This translates to saving over 19 40-hour work weeks and \$1800, assuming 30 seconds and \$0.02 per answer.}
\label{fig_budgetedAnswerDiversityFromCrowd}
\end{figure*} 

Our findings suggest that our Random Forest classifier's overall advantage over our deep learning system arises because of counting questions, as indicated by higher AP scores (\textbf{Figure~\ref{fig_predictiveFeatureAnalysis}}).  For example, the advantage of the initial higher precision (\textbf{Figure~\ref{fig_predictorAnalysis}a}; \texttt{Ours: RF} vs \texttt{Ours: DL}) is also observed for counting questions (\textbf{Figure~\ref{fig_predictiveFeatureAnalysis}b}; \texttt{Ours: RF - Q+I} vs \texttt{Ours: DL - Q+I}).  We hypothesize this advantage arises due to the strength of the Random Forest classifier in pairing the question prior (``How many?") with the image-based SOS features that indicates the number of objects in an image.  Specifically, we expect ``how many" to lead to agreement only for small counting problems.

\section{Capturing Answer Diversity with Less Effort}
\label{sec_resourceAllocation}
We next present a novel resource allocation system for efficiently capturing the \emph{diversity of true answers} for a batch of visual questions.  Today's status quo is to either uniformly collect $N$ answers for every visual question~\cite{AntolAgLuMiBaZiPa15} or collect multiple answers where the number is determined by external crowdsourcing conditions~\cite{BighamJaJiLiMiMiMiTaWhWhYe10}.  Our system instead spends a human budget by predicting the number of answers to collect for each visual question based on whether multiple human answers are predicted to be redundant.  

\subsection*{Answer Collection System}
Suppose we have a budget $B$ which we can allocate to collect extra answers for a subset of visual questions.  Our system automatically decides to which visual questions to allocate the ``extra" answers in order to maximize captured answer diversity for all visual questions.  

The aim of our system is to accrue additional costs and delays from collecting extra answers only when extra responses will provide more information.  Towards this aim, our system involves three steps to collect answers for all $N$ visual questions (\textbf{Figure~\ref{fig_budgetedAnswerDiversityFromCrowd}a}).  First, the system applies our top-performing random forest classifier to every visual question in the batch.  Then, the system ranks the $N$ visual questions based on predicted scores from the classifier, from visual questions most confidently predicted to lead to answer ``agreement" from a crowd to those most confidently predicted to lead to answer ``disagreement" from a crowd.  Finally, the system solicits more ($R$) human answers for the $B$ visual questions predicted to reflect the greatest likelihood for crowd disagreement and fewer ($S$) human answers for the remaining visual questions.  More details below.

\subsection*{Analysis of Answer Collection System}
We now describe our studies to assess the benefit of our allocation system to reduce human effort to capture the diversity of all answers to visual questions.

\paragraph*{Experimental Design}
We evaluate the impact of actively allocating extra human effort to answer visual questions as a function of the available budget of human effort.  Specifically, for a range of budget levels, we compute the total measured answer diversity (as defined below) resulting for the batch of visual questions.  The goal is to capture a large amount of answer diversity with little human effort.

We conduct our studies on the 121,512 test visual questions about real images (i.e., Validation questions 2015 v1.0).  For each visual question, we establish the set of true answers as all unique answers which are observed at least twice in the 10 crowdsourced answers per visual question.  We require agreement by two workers to avoid the possibility that ``careless/spam" answers are treated as ground truth.    

\paragraph*{System Implementation} We collect either the minimum of $S=1$ answer per visual question or the maximum of $R=5$ answers per visual question.  Our number of answers roughly aligns with existing crowd-powered VQA systems, for example with VizWiz, ``On average, participants received 3.3 (SD=1.8) answers for each question"~\cite{BighamJaJiLiMiMiMiTaWhWhYe10}.  Our maximum number of answers also supports the possibility of capturing the maximum of three unique, valid answers typically observed in practice (recall study above).  While more elaborate schemes for distributing responses may be possible, we will show this approach already proves quite effective in our experiments.  We simulate answer collection by randomly selecting answers from the 10 crowd answers per visual question.

\paragraph*{Baselines}
We compare our approach to the following baselines:
\begin{description}
\setlength\itemsep{0em}
\item [\texttt{VQA Algorithm~\cite{AntolAgLuMiBaZiPa15}}:]~\\ As in the previous section, we leverage the output confidence score from the publicly-shared model~\cite{LuLiBaPa15} learned from a LSTM-CNN deep learning architecture to rank the order of priority for visual questions to receive redundancy.
\item [\texttt{Status Quo}:]~\\ The system randomly prioritizes which images receive redundancy.   This predictor illustrates the best a user can achieve today with crowd-powered systems~\cite{BighamJaJiLiMiMiMiTaWhWhYe10,BurtonBrBrNeBiHu12} or with current dataset collection methods~\cite{AntolAgLuMiBaZiPa15,YuPaBeBe15}.
\end{description}

\paragraph*{Evaluation Methodology}  
We quantify the total diversity of answers captured by a resource allocation system for a batch of visual questions $Q$ as follows:

\begin{eqnarray}
D(Q) = \sum_{i=1}^{|B|} |r_i \cap q_i| + \sum_{j=1}^{|Q \setminus B|} |s_j \cap q_j|
\label{eqn_diversityScore}
\end{eqnarray}

\noindent
where $q_{i}$ represents the set of all true answers for the $i$-th visual question, $r_{i}$ represents the set of unique answers captured in the $R$ answers collected for the $i$-th visual question, and $s_{j}$ represents the set of unique answers captured in the $S$ answers collected for the $j$-th visual question.  Given no extra human budget, total diversity comes from the second term which indicates the diversity captured when only $S$ answers are collected for every visual question.  Given a maximum available extra human budget ($B$), total diversity comes from the first term which indicates the diversity captured when $R$ answers are collected for every visual question.  Given a partial extra human budget ($B$), the aim is to have perfect predictions such that the minimum number of answers ($S$) are allocated only for visual questions with one true answer so that all diverse answers are safely captured.  

We measure diversity per visual question as the number of all true answers collected per visual question ($|a \cap b|$).  Larger values reflect greater captured diversity.  The motivation for this measure is to only give total credit to visual questions when all valid, unique human answers are collected.  

\paragraph*{Results}
Our system consistently offers significant gains over today's status quo approach (\textbf{Figure~\ref{fig_budgetedAnswerDiversityFromCrowd}b}).  For example, our system accelerates the collection of 70\% of the diversity by 21\% over the \texttt{Status Quo} baseline.  In addition, our system accelerates the collection of the 82\% of diversity one would observe with VizWiz by 23\% (i.e., average of 3.3 answers per visual question).  In absolute terms, this means eliminating the collection of 92,180 answers with no loss to captured answer diversity.  This translates to eliminating 19 40-hour work weeks and saving over \$1800, assuming workers are paid \$0.02 per answer and take 30 seconds to answer a visual question.  Our approach fills an important gap in the crowdsourcing answer collection literature for targeting the allocation of extra answers only to visual questions where a diversity of answers is expected. 

\textbf{Figure~\ref{fig_budgetedAnswerDiversityFromCrowd}b} also illustrates the advantage of our system over a related VQA algorithm~\cite{AntolAgLuMiBaZiPa15} for our novel application of cost-sensitive answer collection from a crowd.  As observed, relying on an algorithm's confidence in its answer offers a valuable indicator over today's status quo of passively budgeting.  While we acknowledge this method is not intended for our task specifically, it still serves as an important baseline (as discussed above).  We attribute the further performance gains of our prediction system to it directly predicting whether \emph{humans} will disagree rather than predicting a property of a specific \emph{algorithm} (e.g., confidence of the Antol et al. algorithm in its answer prediction).

\section{Conclusions}
\label{sec_conclusions}
We proposed a new problem of predicting whether different people would answer with the same response to the same visual question.  Towards motivating the practical implications for this problem, we analyzed nearly half a million visual questions and demonstrated there is nearly a 50/50 split between visual questions that lead to answer agreement versus disagreement.  We observed that crowd disagreement arose for various types of answers (yes/no, counting, other) for many different reasons.  We next proposed a system that automatically predicts whether a visual question will lead to a single versus multiple answers from a crowd.  Our method outperforms a strong existing VQA system limited to estimating \emph{system} uncertainty rather than \emph{crowd} disagreement.  Finally, we demonstrated how to employ the prediction system to accelerate the collection of diverse answers from a crowd by typically at least 20\% over today's status quo of fixed redundancy allocation. 

\section*{Acknowledgments}
The authors gratefully acknowledge funding from the Office of Naval Research (ONR YIP N00014-12-1-0754) and National Science Foundation (IIS-1065390).  We thank Dinesh Jayaraman, Yu-Chuan Su, Suyog Jain, and Chao-Yeh Chen for their assistance with experiments. 

\bibliography{predictAgreement}
\bibliographystyle{IEEEtran}

\end{document}